\documentclass[letterpaper, 10 pt, conference]{ieeeconf}  

\IEEEoverridecommandlockouts                              

\usepackage{cite}
\usepackage{amsmath,amssymb,amsfonts,esvect}
\usepackage{algorithmic}
\usepackage{graphicx}
\usepackage{textcomp}
\usepackage{xcolor}
\usepackage{romannum}

\title{\LARGE \bf
Design and experimental investigation of a vibro-impact self-propelled capsule robot with orientation control}

\author{Jiajia Zhang, Jiyuan Tian, Dibin Zhu, Yang Liu and Shyam Prasad
\thanks{*This work was supported by the EPSRC under Grant No. EP/R043698/1.}
\thanks{J. Zhang, J. Tian, D. Zhu and Y. Liu are with the College of Engineering, Mathematics and Physical Sciences, University of Exeter, North Park Road, Exeter, UK, EX4 4QF,
        {\tt\small \{jz431,jt535,d.zhu,y.liu2\}@exeter.ac.uk}}%
\thanks{S. Prasad is with the Royal Devon and Exeter NHS Foundation Trust, Barrack Road, Exeter, UK, EX2 5DW, {\tt\small shyamprasad@nhs.net}}%
}

\begin{document}

%
%
%

\maketitle
\thispagestyle{empty}
\pagestyle{empty}

\begin{abstract}
This paper presents a novel design and experimental investigation for a self-propelled capsule robot that can be used for painless colonoscopy during a retrograde progression from the patient's rectum. The steerable robot is driven forward and backward via its internal vibration and impact with orientation control by using an electromagnetic actuator. The actuator contains four sets of coils and a shaft made by permanent magnet. The shaft can be excited linearly in a controllable and tilted angle, so guide the progression orientation of the robot. Two control strategies are studied in this work and compared via simulation and experiment. Extensive results are presented to demonstrate the progression efficiency of the robot and its potential for robotic colonoscopy.
\end{abstract}


\section{Introduction}
Bowel cancer accounts for one third of the total cancer incidence and mortality globally \cite{Lau2020}, and it is the second deadliest cancer \cite{Dceurope2021}, responsible for approximately 170,000 deaths every year in the European Union alone. Detection of bowel cancer and its precursor lesions is currently performed by visual inspection of the colonic mucosa during colonoscopy, which uses a rigid tubular  colonoscope to insert through the patient's rectum. During the procedure, the gastroenterologist pushes the tube gently forward until it reaches the end of the ileum, and then pulls it back and examine the colonic mucosa carefully in real time by using a high-resolution video camera. Although the stiffness of the tube is tunable, patients may experience significant pain during the procedure, which requires a team of clinicians to sedate and monitor patients, and to maintain and decontaminate increasingly complex and expensive devices. Despite major advances in image acquisition and processing over recent decades, the basic design and ergonomics of colonoscopes have barely changed for more than 40 years. The procedure relies on the experience and skills of the gastroenterologists, and lengthy training periods and highly developed professional regulatory frameworks are required. Therefore in colonoscopic practice, there is an urgent need for new modalities that are safe, painless and reliable, which require minimal training for practitioners.

The medical capsule robots \cite{Attanasio2021}, in this case replacing the rigid tube of colonoscopy can be one of the promising solutions to ease patient's discomfort and provide further autonomy to the procedure. A self-propelled colonoscopic robot can use external flexible paddles to anchor its tip \cite{Osawa2020} or the back \cite{Kang2021} of the colonoscope for forward progression. The concept of balloon has been adopted in soft robotics for colonoscopy, such as in \cite{Dehghani2017} the axial expansion of the latex tube between the tip and the anal fixture drives the device forward, and the soft pneumatic inchworm balloon designed in \cite{Manfredi2019} with a combination of two balloons and a soft pneumatic actuator in three degrees of freedom. Another concept uses wires to drag several segments of body of the colonoscope to steer and anchor upon a large bending angle \cite{Bernth2017}. However, all these methods can potentially traumatise the intestinal tissue as anchoring, expanding or reshaping of the colon is required, and manipulating those complex medical devices in the intestinal environment is not an easy task.

\begin{figure}[h!]
\centering
\includegraphics[width=8.3cm]{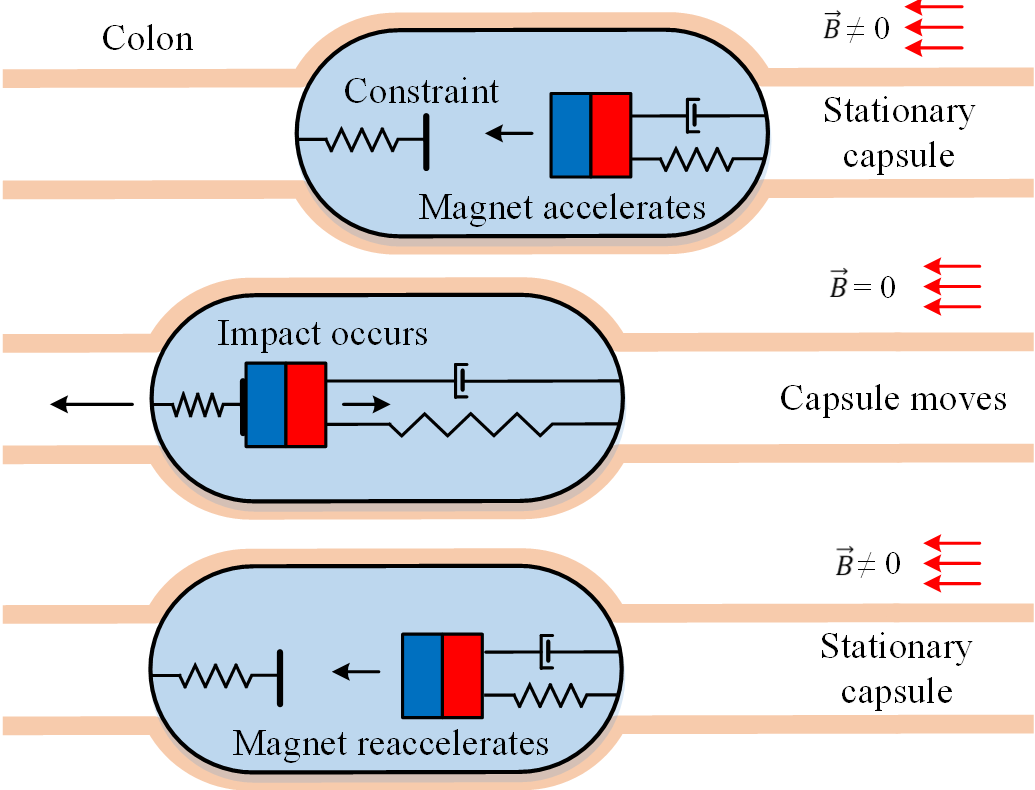}
\caption{Principle of the ``hammer'' capsule.}
\label{concept}
\end{figure}

\begin{figure*}[h!]
\centering
\includegraphics[width=0.75\textwidth]{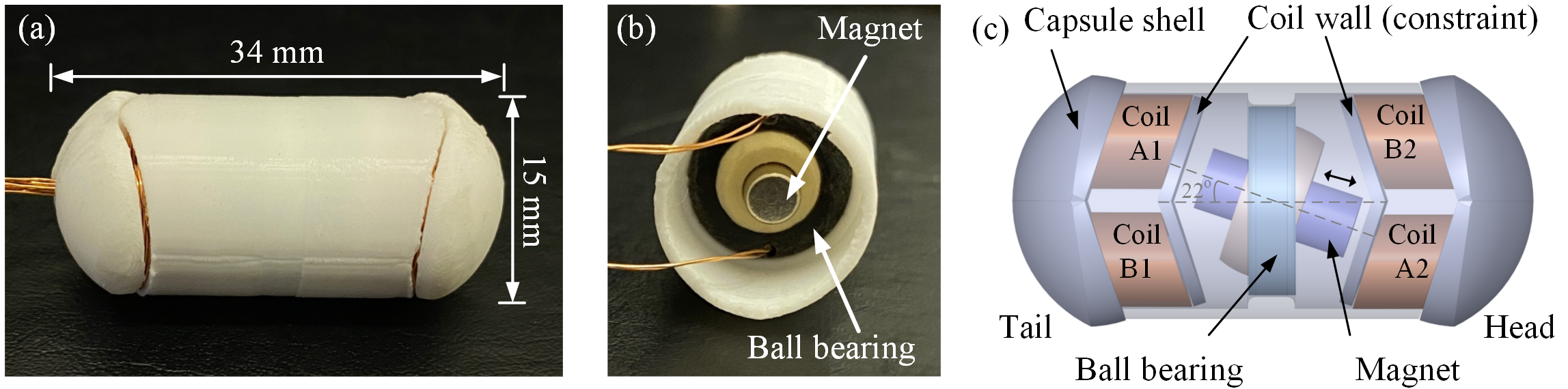}
\caption{(a) External, (b) internal views and (c) schematic of the ``hammer'' capsule prototype.}
\vspace{-10pt}
\label{Solidworks}
\end{figure*}

In this work, we propose a new concept of self-propulsion for colonoscopy by adopting the vibration and impact actuation inside a capsule as illustrated in Fig.~\ref{concept}. This concept was firstly introduced in \cite{Liu20132}, where the principle of the technique is that the rectilinear motion of the capsule can be generated using a periodically driven internal mass interacting with the main body of the capsule as a ``hammer'', in the presence of external resistances. The entire capsule will be progressing at its maximum during the resonance of the ``hammer''. The merit of such a system is its simplicity in mechanical design and control which does not require any external driving accessories, while allowing independent movements in a complex environment. Imagine for example, a self-propelled capsule endoscope entering patient’s rectum directly rather than being swallowed using this method. In this case, many complications and risks induced by powering issues and external propellers, e.g., \cite{Attanasio2021,Osawa2020,Manfredi2019}, can be avoided. The working efficiency of this technique has been demonstrated in different experimental scenarios focusing on rectilinear motion of the capsule. In \cite{Duong2018}, a vibration-driven system with two-sided constraint was studied from numerical simulation to experimental verification. A proof-of-concept prototype of the ``hammer'' capsule in mesoscale was developed and demonstrated experimentally in \cite{Guo2020}. A prototype of the ``hammer'' capsule with 26 mm in length and 11 mm in diameter, aiming for small-bowel endoscopy, was developed and validated in \cite{Liu2020}.

In this study, we extend the concept of the ``hammer'' capsule in one dimension to two dimensional driving, aiming for painless capsule colonoscopy. By adopting this method, vibration and impact forces are transferred from the inner moving mass to the entire capsule system. Different from previous works, the inner mass in the present work is able to change its impact angle, so the capsule is steerable on a 2D plane. The rest of the paper is organised as follows. Section II summarises the requirements of the design and introduces the concept of the driving principle. In Section III, actuation system of the capsule robot is studied in detail. Modelling and simulation of the driving mechanism are studied and compared with experiment in Section IV. Experimental investigation of the prototype under different control parameters is carried out in Section V, and finally, conclusions are drawn in Section VI.

\section{Design Requirement and Concept}

As an invasive tool into patient's rectum, the ``hammer'' capsule should be designed under several constrains, such as dimensions, power and maneuver. As the volume of the robot is strictly restricted while on-board actuation is required, it forms the main challenge of the design. Successful prototyping makes it possible for more integrated implementations e.g., video cameras.

\begin{itemize}
\item Capsule's diameter should be determined based on the diameter of the conventional flexible colonoscope, and its length should be suitable for easy steering in the colon. Considering the the diameter of the standard colonoscope (13 mm) and angulation taking place during the procedure, the dimension of the capsule prototype was determined to be 15 mm in diameter and 34 mm in length.

\item The capsule should be able to steer itself on a 2D plane so to guide itself through sharp angles and narrow passages. Only internal actuation should be considered without any external accessories. To realise the steering function, a new actuation method was applied by adopting four coils, being switched on in different patterns to excite an inner permanent magnet impacting with the front and back constraints. A self-aligning ball bearing was used to guide the permanent magnet to vibrate linearly in different orientations.

\item The safety issues need to be considered. The voltage used in the robot should be as low as possible, so the possibility of exposure to an electric shock could be reduced, and heat dissipation problem can be mitigated accordingly. Although the robot is tethered, the efficiency of its actuation module needs to be optimised, which is mainly influenced by friction and magnetic driving force. As for friction, using a commercial ball bearing can greatly reduce the friction between the vibrated magnet and capsule's body. To optimise the excitation force, the shape of the coils was designed to be elliptic to get the maximised capacity of inner space. Switching on more than one coil simultaneously can generate larger excitation force on the magnet.
\end{itemize}

The photograph and internal views of the prototype are presented in Fig.~\ref{Solidworks}, and its design parameters which were determined based on the previous work \cite{Liu2020} and design requirements summarised above, are listed in Table~\ref{tab:para}. The capsule body was fabricated using polylactic acid via 3D printing. As can be seen from Fig.~\ref{Solidworks}(c), two pairs of identical coils were placed at each side of the capsule with an offset angle. In this way, the neodymium permanent magnet (NdFeB) can vibrate linearly with a maximum angle of 22\textdegree~from the axial direction of the capsule body. This maximum angle was determined by the bearing's ball joint, so can be altered by choosing a proper ball bearing based on the requirement of orientation control.




\begin{table}[h!]
	\caption{Design parameters of the capsule prototype.}\label{tab:para}
	\centering
\begin{tabular}{lcc}
	\hline
	\multicolumn{1}{c}{Parameters}& Units& Values\\
	\hline
	Capsule diameter & mm & 15\\
	Capsule length & mm & 34\\
    Magnet diameter& mm & 4\\
	Magnet length & mm & 10\\
    Magnet weight & g & 0.92\\
    Magnet magnetisation & A/m & $8.38\times10^5$\\
    Coil current & A & 0.5\\
    Coil turns & -- & 50\\
	Coil thickness & mm & 1.3\\
	Ball bearing thickness & mm & 3\\
	Ball bearing friction coefficient & -- & 0.097\\
	Stroke length (total gap)& mm & 2.4\\
    Prototype weight & g & 5.38\\
	\hline
\end{tabular}
\end{table}

\section{Actuation System}
\subsection{Driving Mechanism}\label{Sec:MM}

\begin{figure}[h!]
\centering
\includegraphics[width=8.5cm]{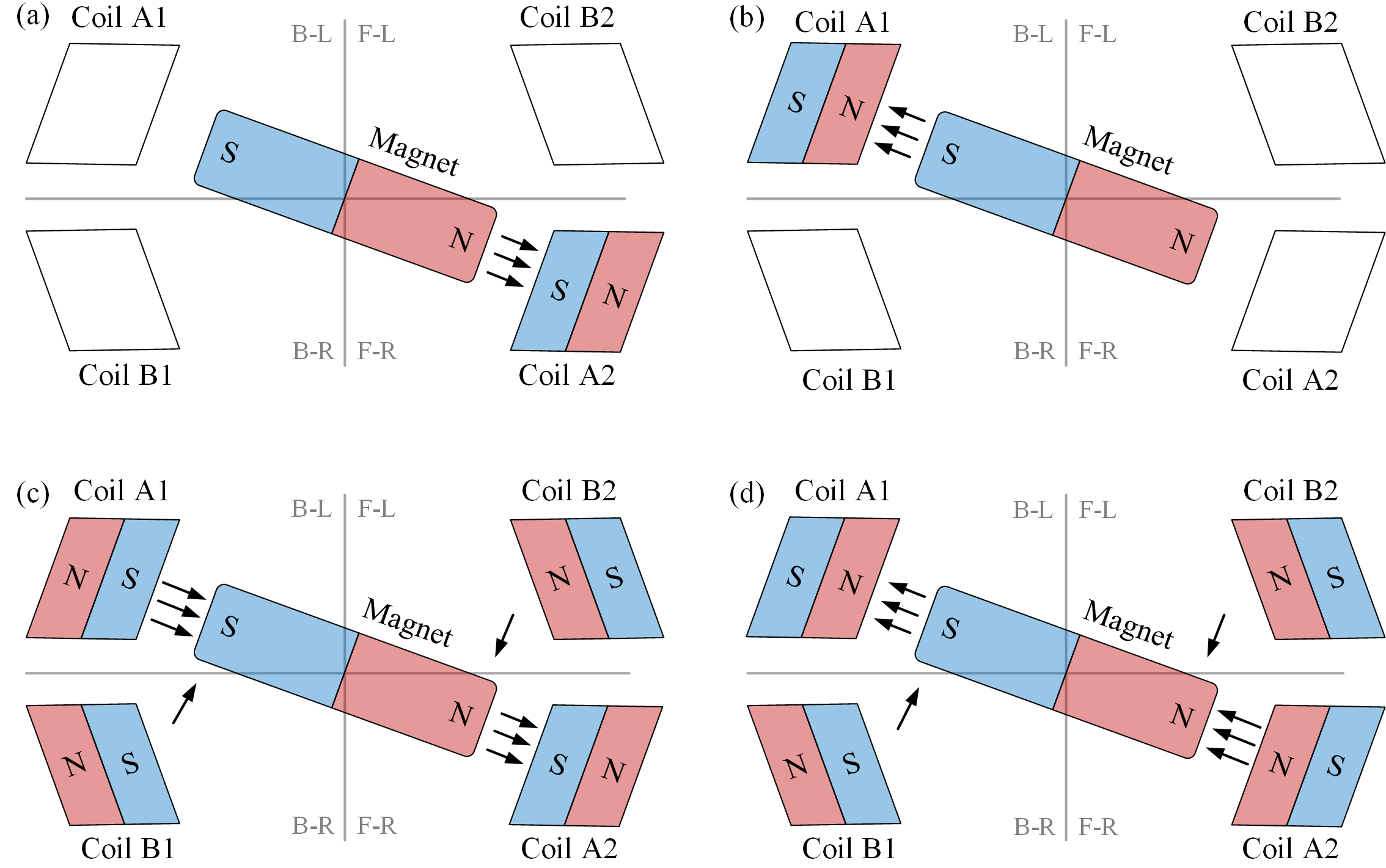}
\caption{The (a,b) one-coil and (c,d) four-coil driving methods.}
\label{Prin02}
\end{figure}

In this work, four coils, A1, A2, B1 and B2, were assembled for directional control of the prototype as presented in Fig.~\ref{Solidworks}(c), where A1-A2 forms a driving pair and B1-B2 forms another one to control the vibrational direction of the permanent magnet through the ball bearing. Here, we propose two control methods, one-coil and four-coil excitations, for driving the magnet. The driving principles for these two methods are illustrated in Fig.~\ref{Prin02}, where the square-wave (on-off) signal is applied. As can be seen from Figs.~\ref{Prin02}(a) and (b), for the one-coil method, in first period of excitation, A2 is switched on and A1 is off, while in second period of excitation, A2 is off and A1 is switched on. By repeating this excitation, the magnet will be attracted by A2 and A1 in turn and makes forward and backward impacts on the constraints (coil wall), while the other two coils, B1 and B2, remain switched off. By adjusting the ratio of the first and second periods of excitation or the frequency of the square-wave signal, the entire capsule may overcome its external friction to progress either in forward-right (F-R) or backward-left (B-L) direction.

For the four-coil method, all four coils are working simultaneously. In the first period of excitation, A1 and A2 are switched on but generating magnetic field in opposite directions, so the magnet can be attracted by A2 and repelled by A1, while B1 and B2 also provide repelling magnetic forces. In the second period of excitation, A1 and A2 reverse the directions of their magnetic fields, while B1 and B2 retain their magnetic fields, so the magnet can be attracted by A1 and repelled by A2. By adjusting the ratio of the first and second periods of excitation or the frequency of the square-wave signal, the entire capsule may overcome the external friction to progress either in F-R or B-L direction. By using the same driving principle, the capsule prototype can also be driven in forward-left (F-L) or backward-right (B-R) direction. Thus, the prototype can be controlled on a 2D plane. It is worth noting that the tilted angle of the magnet is fixed at 22\textdegree~for the one-coil method, but is changeable for the four-coil method by adjusting the repelling force. For example, as illustrated in Figs.~\ref{Prin02}(c) and (d), magnet's tilted angle can be changed for orientation control by adjusting the voltage applied to the repelling coil pair, B1 and B2.

\subsection{Modelling of the Magnetic Force}\label{Sec:MP}
To calibrate the magnetic force and torque acting on the magnet from the coils, Biot–Savart's law \cite{Gellert2013} was employed to calculate the magnetic field as
\begin{equation} \label{eq:Biot0}
\vv{B}=\frac{\mu_0}{4\pi}\int_{}^{}\frac{I\cdot\mathrm{d}\vv{l}\times \vv{r}}{r^3},
\end{equation}
where $I$ is the current, d$\vv{l}$ is a vector, whose magnitude is the length of the differential element of the wire in the direction of the current, $\vv{B}$ is the net magnetic field, $\mu_0$ is the magnetic constant, $\vv{r}$ is the displacement vector in the direction pointing from the wire element towards the point at where the field is calculated.

The analytical solution of magnetic field generated by an ellipse coil can be derived by breaking down the ellipse shape into multi-polygons and making linear superposition calculations. The force and torque on the magnet can be calculated by using \cite{Schuerle2013}
\begin{equation} \label{eq:force}
\bold{F} = v (\bold{M} \cdot\nabla) \bold{B} =v [\frac{\partial \bold{B}_x}{\partial x} \frac{\partial \bold{B}_y}{\partial y} \frac{\partial \bold{B}_z}{\partial z}]^{T} \bold{M}
\end{equation}
\begin{equation} \label{eq:torque}
\bold{T} = v \bold{M} \times \bold{B}
\end{equation}
where $\bold{M}$ is the magnetisation of the magnet, $v$ is the volume of the magnet, $\nabla\bold{B}$ is the gradient of the magnetic field and $\bold{B}$ is the magnetic field which is derived from Eq.~\eqref{eq:Biot0}. The force tends to drag or repel the magnetisation vector to where higher magnetic field exists, while the torque tends to align the magnetisation vector to the magnetic field applied.

\section{Simulation and Verification}\label{MM}
\begin{figure}[h!]
\centering
\includegraphics[width=8.5cm]{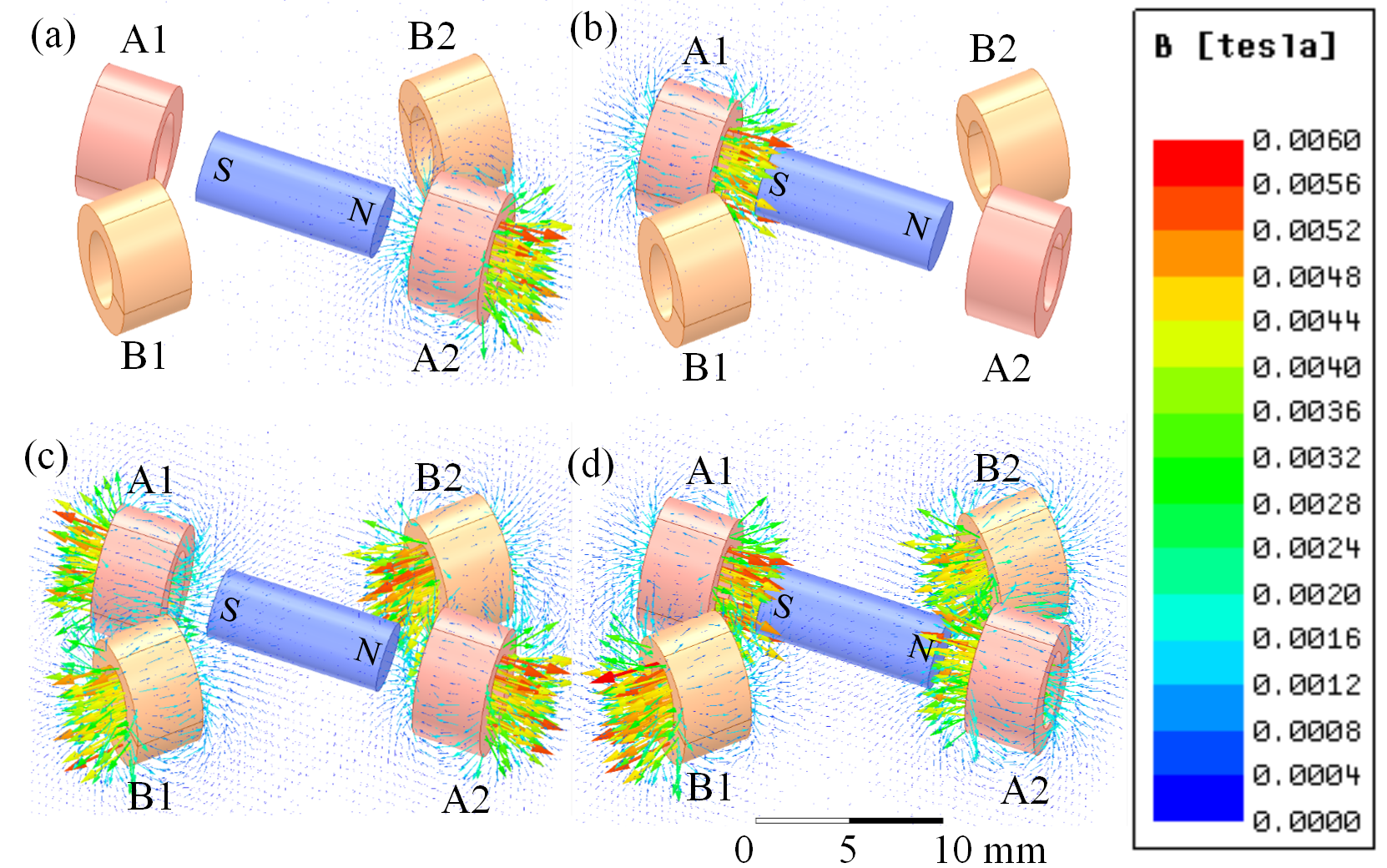}
\caption{Simulation results of the (a,b) one-coil and (c,d) four-coil driving methods calculated using ANSYS Maxwell.}
\label{Simulation}
\end{figure}

A finite element model of actuation system of the prototype was built by using ANSYS Maxwell, where the dimensions of the model are listed in Table~\ref{tab:para}. According to the simulation, the distributions of the magnetic fields by using the two proposed driving methods are plotted in Fig.~\ref{Simulation}, which are in good agreement with the driving principle illustrated in Fig.~\ref{Prin02}. To better display the strength of the actuation coils, the magnetic fields generated by the coils are presented in these figures, and the magnetic field of the magnet is hidden.

\begin{figure}[h!]
\centering
\includegraphics[width=8.5cm]{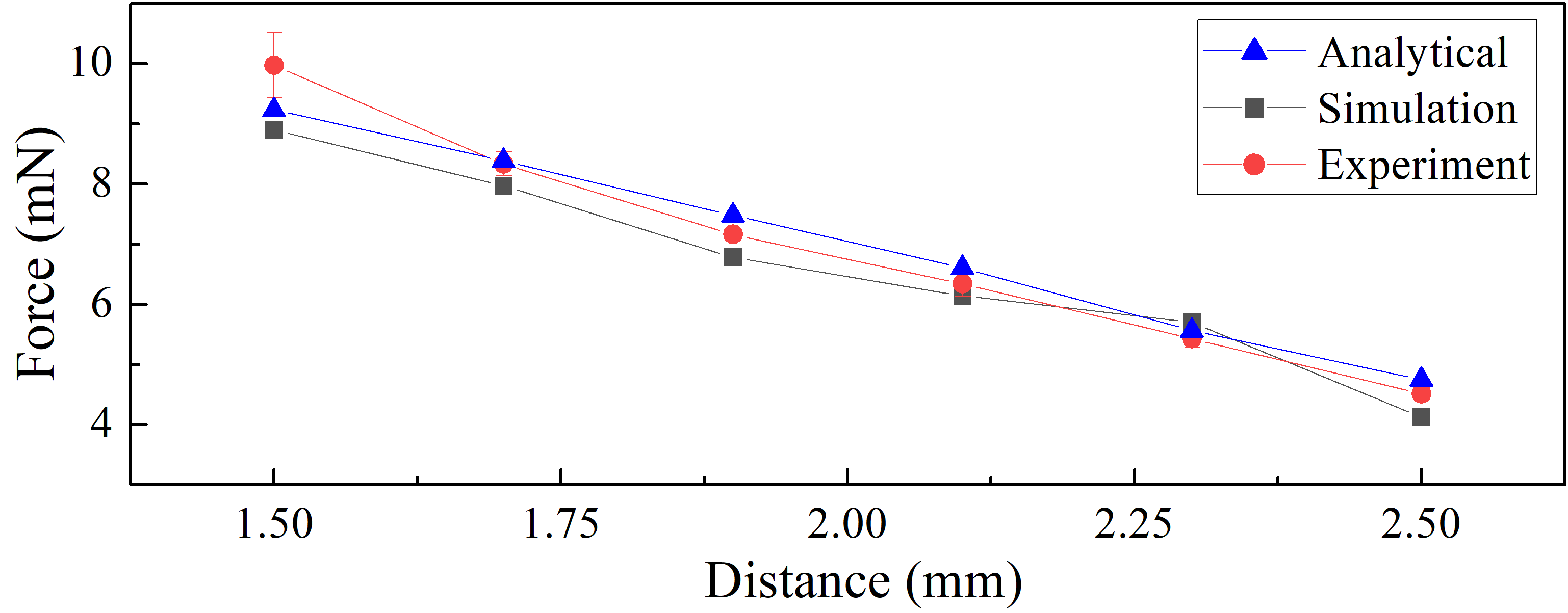}
\caption{Excitation forces on the magnet versus the distances between the magnet and the coil obtained by the analytical solution (blue-triangle) using Eq.~\eqref{eq:force}, simulation (black-square) using ANSYS Maxwell and experiment (red-dot).}
\label{exp}
\end{figure}

Figure~\ref{exp} presents the forces generated on the magnet from one coil excitation with a varying distance between the magnet and the coil, where blue-triangle, black-square and red-dot lines denote the analytical using Eq.~\eqref{eq:force}, simulation using ANSYS Maxwell and experimental results, respectively. The experimental force was acquired directly by weighing the magnet using a digital weight while placing the coil vertically above the magnet at different distances, and an averaged result was calculated from the readings of the weight by repeating the measurement for five times. Finally, the weight result was multiplied by 9.81 N/kg to get the magnetic force generated from the coil.

\section{Experiments and Discussion}\label{ED}
\begin{figure}[h!]
\centering
\includegraphics[width=8.3cm]{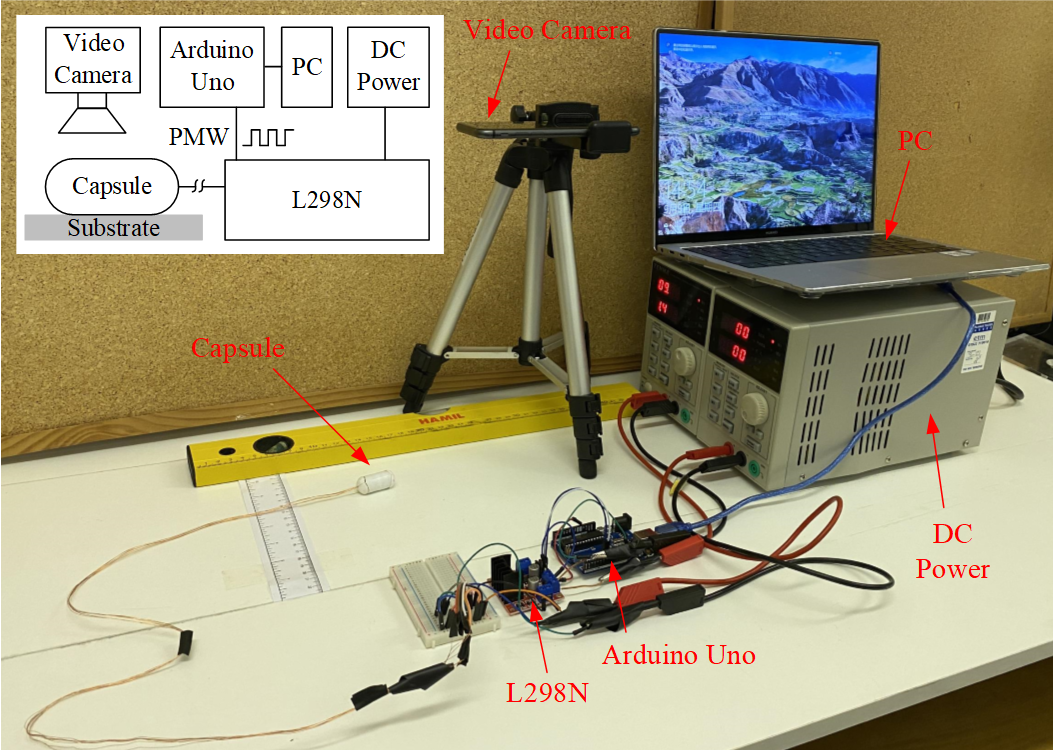}
\caption{Photograph and schematic of the experimental setup.}
\label{Control}
\end{figure}

The photograph and schematic of the experimental setup are presented in Fig.~\ref{Control}, where a video camera was used to record the movement of the capsule. To excite the permanent magnet, an AC driving current was supplied to the coils to create a constantly alternating magnetic field. Thus, a pulse-width-modulation (PWM) signal was generated using Arduino Uno. The Arduino board sent control signal to a H-bridge motor controller, L298N, powered by an external power supply. The dual H-bridge configuration of L298N was able to support both the one- and four-coil driving methods. The frequency and duty cycle of the PWM signal can be adjusted, while the current from the power supply was kept constant during the experiment. The duty cycle of the driving signal is the proportion of the positive excitation duration in one driving period.

\subsection{One-Coil Driving Method}
The schematic diagrams of the control circuits for the one- and four-coil driving methods are presented in Fig.~\ref{pins}. For the one-coil driving method, the driving coils A1 and A2 were connected to OUT1 and OUT2 of L298N. Each coil was connected to a diode in series as shown in Fig.~\ref{pins}(a). These diodes limited the current direction flown into the coils. In the first period of excitation, A1 was excited to attract the magnet to impact the front constraint. The diode in the A2 branch blocked the current so that it was not excited. In the second period of excitation, A2 was excited to attract the magnet to impact the back constraint, while A1 was not excited as the diode in its branch blocked the current.

\subsection{Four-Coil Driving Method}
\begin{figure}[h!]
\centering
\vspace{+10pt}
\includegraphics[width=8.0cm]{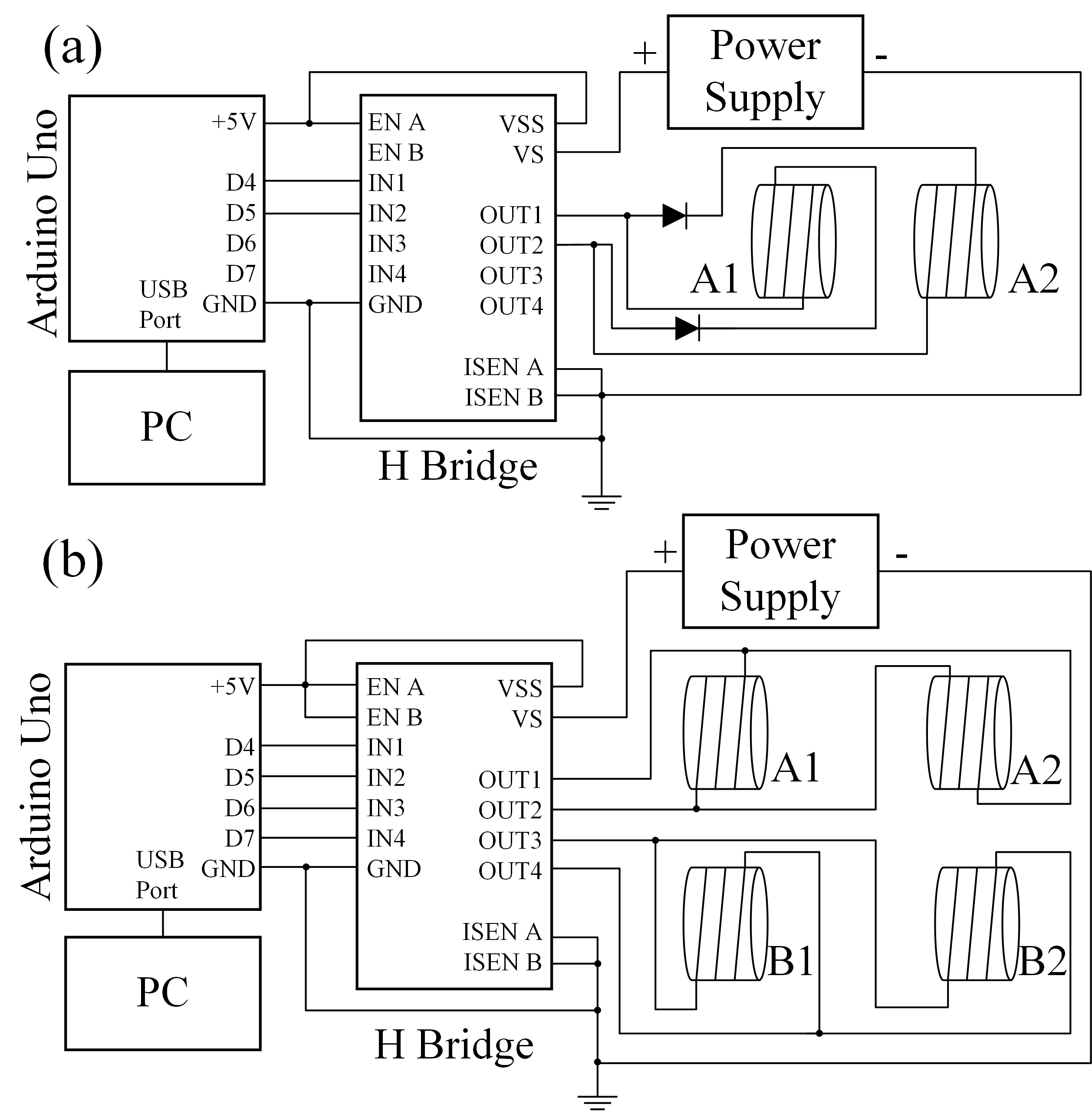}
\caption{Schematic diagrams of the control circuits for the (a) one-coil and (b) four-coil driving methods.}
\label{pins}
\end{figure}

As shown in Fig.~\ref{pins}(b), for the four-coil driving method, the driving coils A1 and A2 were connected to L298N's OUT1 and OUT2 in an opposite switching current flow direction to generate two opposite magnetic fields by using one H-bridge circuit. When coil A1 was supplied with a positive voltage, coil A2 was supplied with a negative voltage. Under different frequencies and duty cycles, the constantly alternating magnetic field can drive the magnet inside the capsule leading to the impacts with the front and back constraints. In the meanwhile, auxiliary coils B1 and B2 were connected to OUT3 and OUT4 with the same constant current flow direction. In this case, the auxiliary coils can provide a repelling torque on the magnet so enhancing its linear vibration along a desired tilted direction.

\subsection{Experimental Results}
\begin{figure}[h!]
\centering
\vspace{+10pt}
\includegraphics[width=8.5cm]{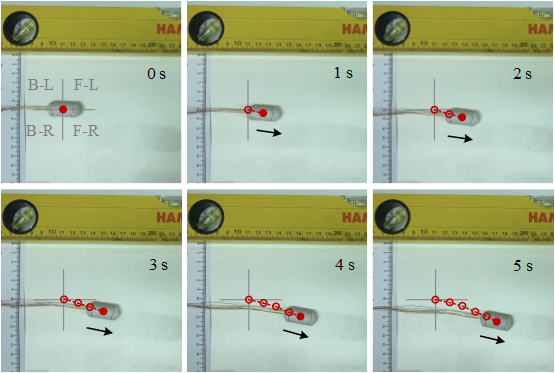}
\caption{Capsule's trajectory on a smooth surface by using the four-coil driving method.}
\label{plane}
\end{figure}

The capsule's movement on a smooth surface in 5 seconds was recorded by the video camera and displayed in Fig.~\ref{plane}, where the trajectory of the capsule was denoted by the red circle line and the black arrows indicated the movement direction of the capsule. In this test, the four-coil driving method was applied, and the prototype was operated at the frequency of 30 Hz and the duty cycle of 60\%. As can be observed from the figure, the capsule moved towards the F-R direction on the plane. According to our measurement, the trajectory of the capsule had a deviated angle of 22\textdegree~from its original coordinate which was consistent with our design.

\begin{figure}[h!]
\centering
\includegraphics[width=8.5cm]{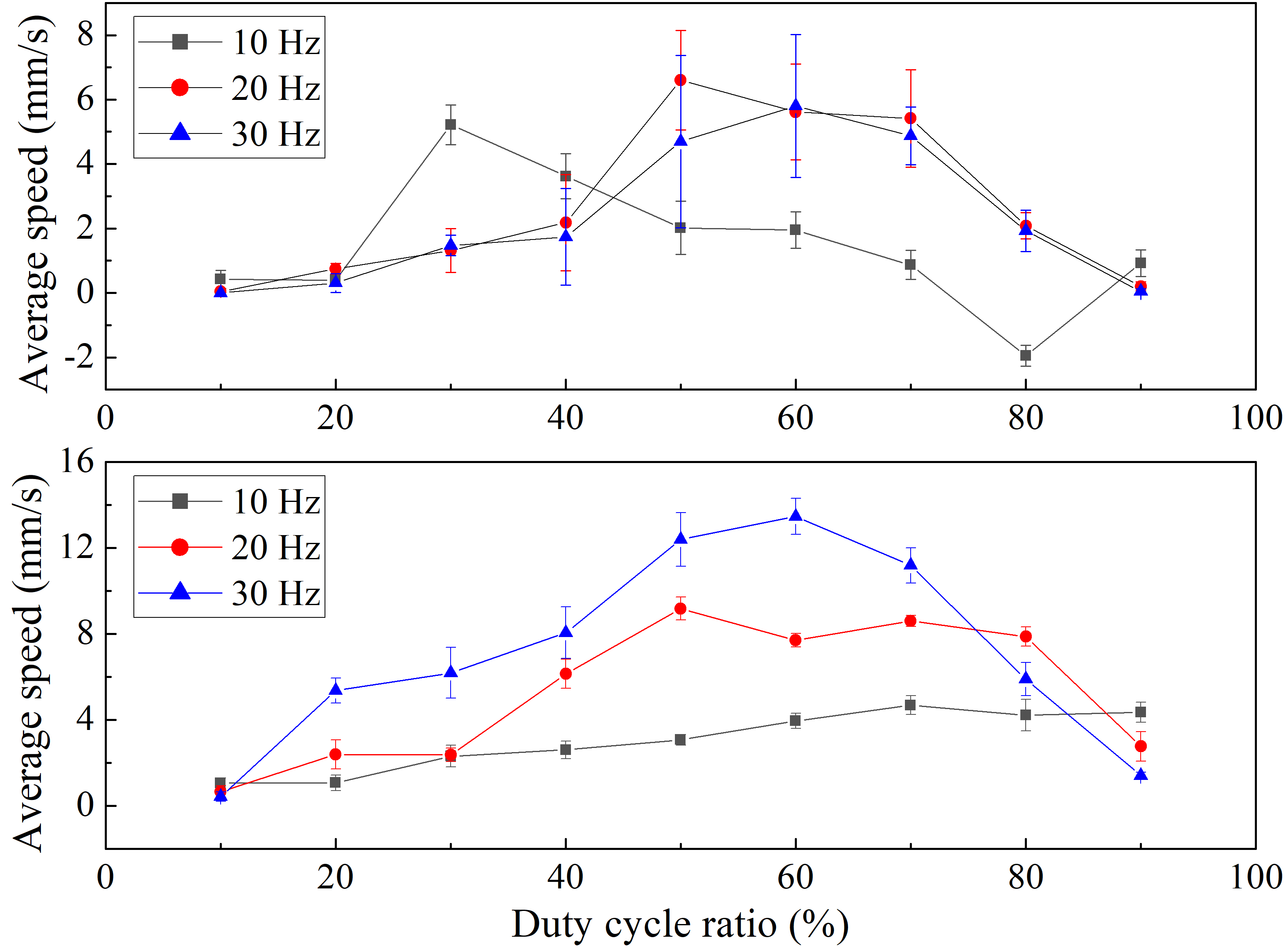}
\caption{Experimental results obtained by using the (a) one-coil and (b) four-coil driving methods on a smooth surface under different frequencies and duty cycles of the excitation showing the standard derivations calculated from five repeated tests for each experiment.}
\label{Experimentdata}
\end{figure}

To optimise the speed of the prototype, both the one-coil and four-coil driving methods were tested on the smooth surface under different combinations of frequencies and duty cycles. The range of the frequency was selected from [10, 30] Hz since the capsule can move efficiently within this range according to the studies in \cite{Guo2020,Liu2020}. For each test, it was repeated fives times, and its averaged speed was plotted in Fig.~\ref{Experimentdata}.

\begin{figure}[h!]
\centering
\includegraphics[width=8.5cm]{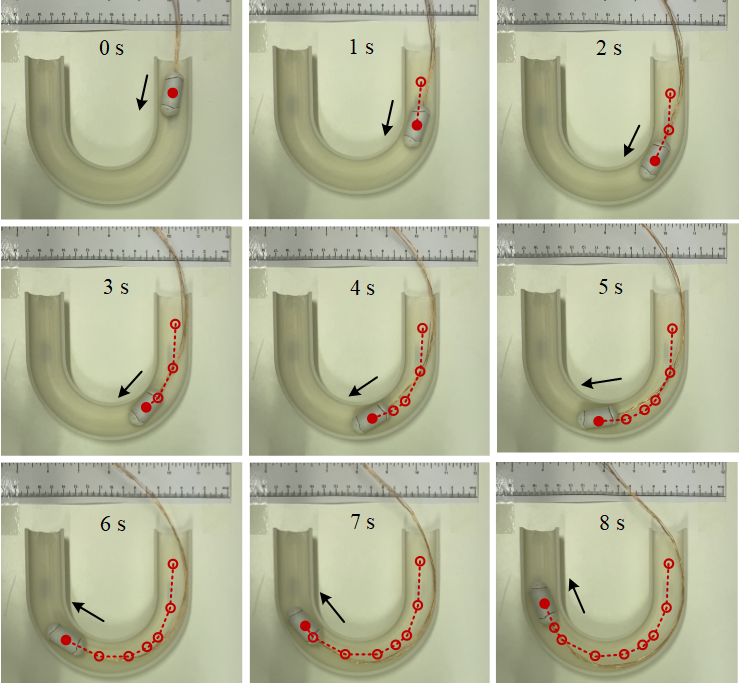}
\caption{Capsule's trajectory on a curved tube by using the four-coil driving method.}
\label{curve}
\end{figure}

\begin{figure}
\centering
\includegraphics[width=7.3cm]{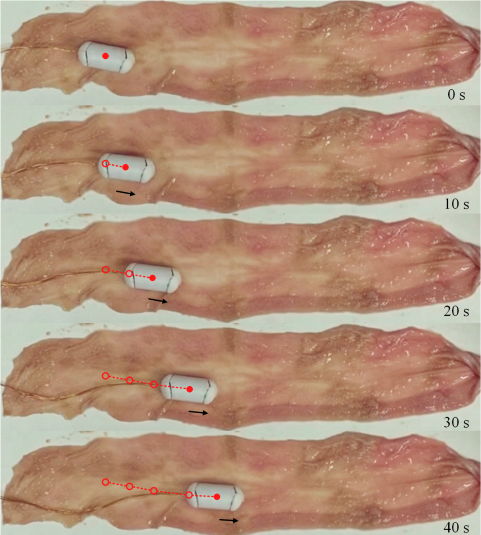}
\caption{Capsule's trajectory on a porcine intestine tissue using the four-coil driving method.}
\label{OnIntestine}
\end{figure}

To demonstrate the capability of the capsule travelling along a semicircular track, another experiment was carried out by using the four-coil driving method as shown in Fig.~\ref{curve}, where a 8-second trajectory of the capsule marked by the red circle line was presented and the movement direction of the capsule was indicated by the black arrows. The prototype was operated at the frequency of 30 Hz and the duty cycle of 60\%. The track was designed in a total length of 250 mm and a semi-circled curve with the diameter of 100 mm. The inner diameter of the track was 25 mm, and it was 3D printed by polylactic acid. During the experiment, the track was fixed on the testing bench, and the test was repeated ten times with video recorded to calculate the averaged time of the passage. As the averaged passage time was $16.1$ s, the average movement speed of the capsule in the semicircular track was $15.53$ mm/s. To validate the efficiency of the prototype, Fig.~\ref{OnIntestine} presents the experimental result of the capsule moving on a cut-open porcine intestine tissue. The test was also operated at the frequency of 30 Hz and the duty cycle of 60\% indicating an average speed of 2.66 mm/s.

According to the experimental results on the plane as shown in Fig.~\ref{Experimentdata}, most of the scenarios were forward progression, while a steady backward progression was obtained by the one-coil method operated at the frequency of 10 Hz and the duty cycle of 80\%. Next, the efficiencies of movement using the one-coil and four-coil methods under the same frequency and duty cycle were compared in Table~\ref{tab:result}. The trend of the experimental results indicates that the four-coil method is more effective at high frequencies (20 and 30 Hz) with a duty cycle around 50--70\%, while the one-coil one has a better performance at low frequency (10 Hz) with both forward and backward movement.


Based on the video observations shown in Fig.~\ref{plane}, the deviated angle of progression was not influenced by the frequency or the duty cycle, but by the tilted angle of the magnet. The video records show that the deviated angles of movement have larger differences for the repeated tests at low speed movement compared to those at high speed. The averaged deviated angle of progression for the one-coil and four-coil methods are 22.54\textdegree~and 23.65\textdegree~, respectively. These results are in good agreement with the design, so the direction of movement of the capsule can be controlled by the tilted angle of the magnet. In addition, temperature change was measured during the experiment for the four-coil method in which all coils were activated and the maximum heat was generated. The records by using an infrared sensor for 10 minutes with three repeats showed that the temperature of the coils rose from the room temperature 18.3\textdegree~C to a stable one around 25\textdegree~C after six minutes of running.

\begin{table}[h!]
	\caption{Maximum average speeds under different control parameters.}\label{tab:result}
	\centering
\begin{tabular}{cccc}
	\hline
	Method & Frequency (Hz) & Duty cycle (\%)& Speed (mm/s)\\
	\hline
        & 10 & 30& $5.21$\\

    One-coil & 20 & 50& $6.60$\\

        & 30 & 60& $5.80$\\
	\hline
	    & 10 & 70& $4.69$\\

    Four-coil & 20 & 50& $9.18$\\

        & 30 & 60& $13.47$\\
	\hline
\end{tabular}
\end{table}

\section{Conclusions}\label{C}
In this paper, we presented a novel design by using the electromagnetic vibration and impact to propel a millimetre-scale capsule prototype for the potential of painless colonoscopy. The prototype was designed according to the clinical requirements that can be scaled up to include a camera unit, and two driving methods were proposed and experimentally tested. The electromagnetic actuation system of the prototype was modelled using finite element analysis, and the force and torque generated by the actuation module inside the capsule were verified. Based on the experimental results, the one-coil driving method can propel the capsule either forward or backward at a low progression speed, while only forward progression at a high speed was observed for the four-coil driving method. In addition, the four-coil driving method was successfully demonstrated on a smooth plane, a curved track and a cut-open porcine intestine tissue. Due to its light-weight body, the tether may have a negative effect on the capsule's dynamics, which can be mitigated by scaling up the prototype with a larger payload. The friction in the real colon should be smaller than the dry surface, but the anatomy of the colon is more complex, e.g., the haustra of the colon, so its resistance is larger. The study in \cite{Liu2020b} verified that the propulsive force generated by the self-propelled capsule can overcome the intestinal resistance, but will not cause any intestinal trauma at a concerned level \cite{Tian2021}. Future work will focus on the design optimisation for the prototype and its tests using a colon phantom.



\end{document}